\documentclass[10pt,twocolumn,letterpaper]{article}

\usepackage[pagenumbers]{iccv} 

\usepackage{multirow}
\usepackage{indentfirst}
\usepackage{graphicx}
\usepackage{blindtext}
\usepackage{caption}
\usepackage{marvosym}
\usepackage[accsupp]{axessibility} 

\definecolor{iccvblue}{rgb}{0.21,0.49,0.74}
\usepackage[pagebackref,breaklinks,colorlinks,allcolors=iccvblue]{hyperref}
\usepackage{pifont}
\newcommand{\cmark}{\ding{51}}%
\newcommand{\xmark}{\ding{55}}%

\def\paperID{5506} 
\def\confName{ICCV}
\def\confYear{2025}

\title{Democratizing High-Fidelity Co-Speech Gesture Video Generation}

\author{
  Xu Yang$^{1*\ddagger}$ \quad
  Shaoli Huang$^{2*}$ \quad
  Shenbo Xie$^{1*}$ \quad
  Xuelin Chen$^2$ \quad
  Yifei Liu$^1$ \quad
  Changxing Ding$^{1\dagger}$ \\
  $^1$South China University of Technology \quad
  $^2$Tencent AI Lab \quad \\
  {\tt\small ftyang\_xu@mail.scut.edu.cn} \quad
  {\tt\small shaol.huang@gmail.com} \quad
  {\tt\small eeshbx34@mail.scut.edu.cn} \\ 
  {\tt\small xuelin.chen.3d@gmail.com} \quad
  {\tt\small ft\_lyf@mail.scut.edu.cn} \quad
  {\tt\small chxding@scut.edu.cn} 
}

\begin{document}
\twocolumn[{%
\renewcommand\twocolumn[1][]{#1}%
\maketitle
\begin{center}
    \centering
    \captionsetup{type=figure}
    \includegraphics[width=1\textwidth]{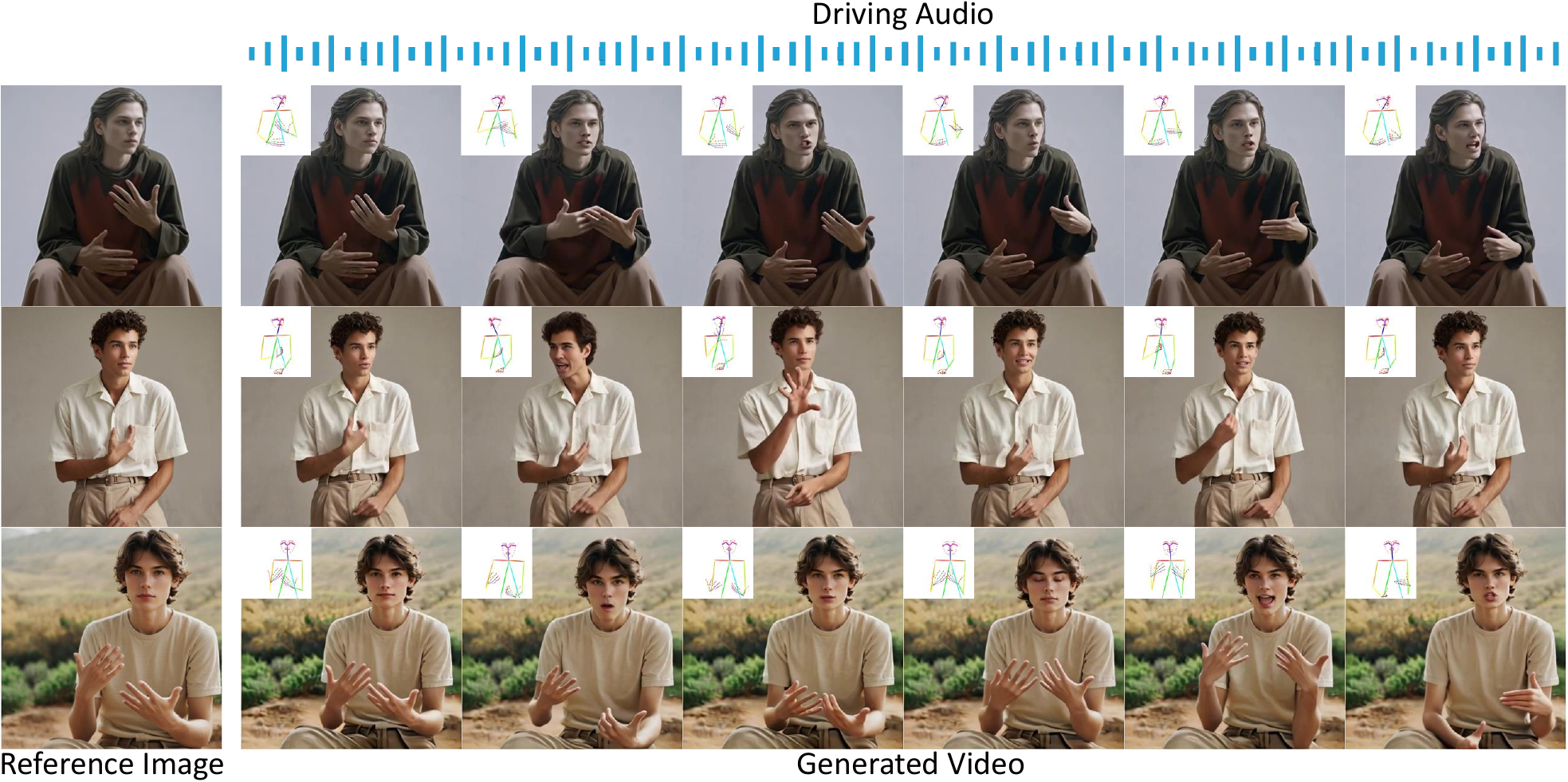}
    \vspace{-7mm}
    \captionof{figure}{Examples of co-speech gesture videos created by our framework. These examples demonstrate the advantages of our framework regarding visual quality and synchronization while being adaptable to various speakers and contexts. To address privacy concerns, all reference images of speakers are synthesized using AI.}
    \label{fig:teaser}
\end{center}%
}]

\makeatletter
\begingroup
\renewcommand*{\@makefnmark}{}
\footnotetext{$^*$Equal Contribution.}
\footnotetext{$^\dagger$Corresponding Author.}
\footnotetext{$^\ddagger$Part of his work was done during an internship at Tencent AI Lab.}
\renewcommand*{\@makefnmark}{\hss\@textsuperscript{\normalfont\@thefnmark}}
\endgroup
\makeatother

\vspace{-5mm}
\begin{abstract}
Co-speech gesture video generation aims to synthesize realistic, audio-aligned videos of speakers, complete with synchronized facial expressions and body gestures. This task presents challenges due to the significant one-to-many mapping between audio and visual content, further complicated by the scarcity of large-scale public datasets and high computational demands. We propose a lightweight framework that utilizes 2D full-body skeletons as an efficient auxiliary condition to bridge audio signals with visual outputs. Our approach introduces a diffusion model conditioned on fine-grained audio segments and a skeleton extracted from the speaker's reference image, predicting skeletal motions through skeleton-audio feature fusion to ensure strict audio coordination and body shape consistency. The generated skeletons are then fed into an off-the-shelf human video generation model with the speaker's reference image to synthesize high-fidelity videos. To democratize research, we present CSG-405—the first public dataset with 405 hours of high-resolution videos across 71 speech types, annotated with 2D skeletons and diverse speaker demographics. Experiments show that our method exceeds state-of-the-art approaches in visual quality and synchronization while generalizing across speakers and contexts. Code, models, and CSG-405 are publicly released at \href{https://mpi-lab.github.io/Democratizing-CSG/}{%
    \url{https://mpi-lab.github.io/Democratizing-CSG/}%
}
\end{abstract}  
\vspace{-6mm}
\section{Introduction}
\label{sec:intro}
\interfootnotelinepenalty=10000
\begin{table*}[t]
    \centering
    \tiny  
    \renewcommand{\arraystretch}{0.85}  
    \caption{Comparison in statistics between our CSG-405 database and existing public ones for co-speech gesture video generation.}
    \resizebox{1\linewidth}{!}{%
        \begin{tabular}{l|ccccc|ccc}
            \toprule
            & \multicolumn{5}{c|}{\text{Video}} & \multicolumn{3}{c}{\text{Motion Annotation}} \\ 
            \multirow{-2}{*}{\text{Dataset}} & \#Clips & \#Hours & Resolution & \#Speakers & \#Speech Types & Face & Body & Hands \\ 
            \midrule
            PATS~\cite{pat2} & 4,800 & 13.1 & 256$\times$256 & 4 & 2 & \xmark & \cmark & \cmark \\ 
            TED-talks~\cite{ted} & 1,322 & 3.1 & 384$\times$384 & 391 & 1 & \xmark & \xmark & \xmark \\ 
            CSG-405 & 147,550 & 405 & 512$\times$512 & 17787 & 71 & \cmark & \cmark & \cmark \\ 
            \bottomrule
        \end{tabular}%
    }

    \label{tab:dataset comp} 
\end{table*}

\begin{figure*}[h]
    \centering
    \includegraphics[width=\linewidth]{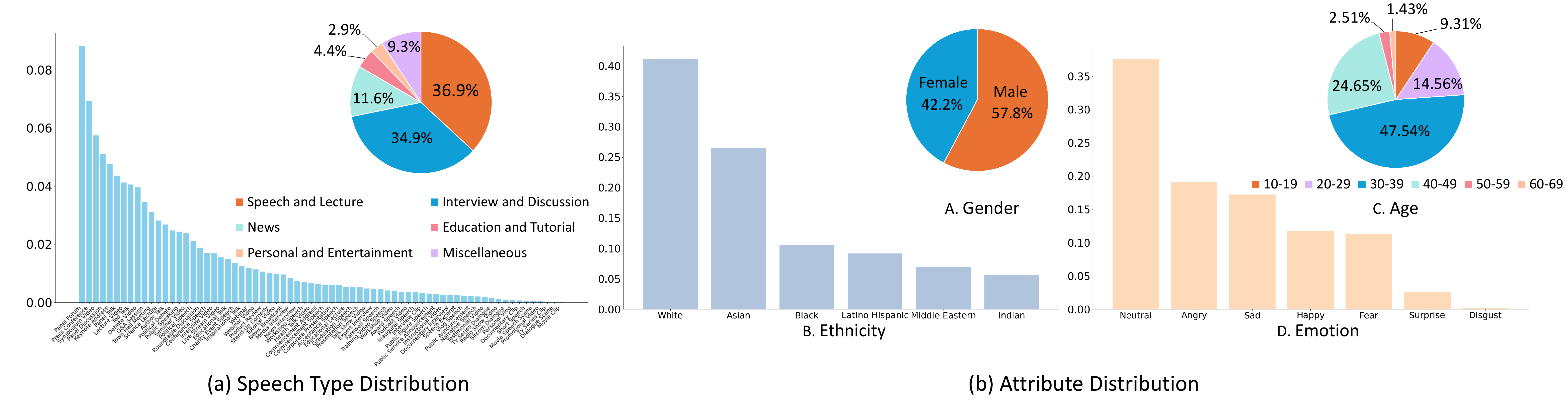}
    \vspace{-4mm}
    \caption{More details of CSG-405. (a) The proportion of clips for each speech type. (b) Attribute distribution in gender, ethnicity, age, and emotion.}
    \label{fig:keypoints and dataset attribute}
    \vspace{-6mm}
\end{figure*}

Co-speech gesture video generation aims to synthesize a realistic video of a person speaking, and the obtained video has to be aligned with both the given speech audio and the reference image of the speaker. It has broad applications, including human-machine interaction~\cite{S2G-MDDiffusion} and digital entertainment~\cite{angie}. Notably, this task presents unique challenges compared to talking face generation~\cite{echomimic,emo,follow-your-emoji,letstalk} and full-body human video generation~\cite{animateanyone,magicanimate,champ}. Compared to the former, it must produce videos that include both facial expressions and hand gestures. Compared to the latter, it must generate audio-aligned talking humans rather than general human movements like dancing.


To synthesize a co-speech gesture video, early methods~\cite{angie,S2G-MDDiffusion,diffted,vlogger} for this task typically involve warping the reference image of the speaker based on the audio condition. However, the warping operation tends to result in artifacts  when encountering significant pose variations. Recent works~\cite{vlogger,cyberhost,echomimicv2} utilize diffusion models~\cite{ddpm,stablediffusion} to achieve warping-free synthesis. Moreover, to facilitate the coordination between visual content and the audio, they impose auxiliary control conditions, e.g., 3D human motions~\cite{vlogger} and hand skeletons extracted from a reference video~\cite{echomimicv2}, on the diffusion model.

Despite these efforts, high-fidelity co-speech gesture video generation is still challenging due to the substantial one-to-many mapping between the audio condition and visual contents~\cite{vlogger,cyberhost}. Recent efforts to handle this problem have mainly been made by industry~\cite{vlogger,cyberhost,echomimicv2}. Their common ground is collecting large-scale private training data, e.g., 2,200 hours of video data in~\cite{vlogger} and 200 hours in~\cite{cyberhost}, and train their models using high-end GPU devices to bridge this mapping. In comparison, existing publicly available datasets for this task are small in size and lack diversity in speech scenarios.   For example, existing works~\cite{angie,S2G-MDDiffusion,diffted} only adopt the 13 hours of video data selected from the PATS dataset~\cite{pat2} and 3 hours of video data from TED talks in the TED-talks dataset~\cite{ted} for experiments. This situation hampers democratization in the research of high-fidelity co-speech gesture video generation.

Our key insight here is that the sequence of expressive 2D full-body skeletons is an essential and economical auxiliary condition to relieve the above one-to-many mapping problem. On the one hand, the 2D skeleton is much more concise than visual data, and its synthesis is computationally efficient. On the other hand, the 2D full-body skeleton has been a popular auxiliary condition for existing diffusion-based human video generation models~\cite{animateanyone,mimicmotion,echomimicv2,stableanimator}, meaning we can easily enhance their ability to generate co-speech gesture videos with improved skeleton conditions. However, high-fidelity audio-to-skeleton prediction is non-trivial, as any artifact, e.g., misalignment between the audio and lip movements, significantly harms the quality of obtained speech videos. To handle this problem, we adopt two strategies. First, we condition this prediction with the skeleton extracted from the reference image of the target speaker, which enables the generated skeletons to conform to the speaker’s body shape. Second, we concatenate the embeddings of skeletons and those of audio segments along the feature dimension as the input of the diffusion model, which enforces strict coordination compared with other conditioning strategies, e.g., cross-attention~\cite{transformer}. Finally, we feed the obtained skeleton sequence, along with the reference image of the speaker, into an off-the-shelf human video generation model to synthesize co-speech gesture videos.

Moreover, we contribute a large-scale dataset named Co-Speech Gesture (CSG-405) that is publicly available\footnote{We release all links, tools, codes, and skeleton data for downloading and automatically preprocessing the involved raw speech videos.}. As shown in Table~\ref{tab:dataset comp}, CSG-405 contains 147,550 clips, covering 71 common speech types, totaling 405 hours. Specifically, each original video has passed a multi-stage filtering process that includes skeleton-detection quality and audio-lip sync assessment. These videos are then segmented into clips of 5-15 seconds, followed by cropping and resizing to a resolution of 512$\times$512 pixels. Compared to existing public datasets~\cite{ted,pat2}, our dataset features a large number of speakers, diverse speech scenarios, rich 2D skeleton annotations, and high image resolution. As shown in Figure~\ref{fig:keypoints and dataset attribute}, it includes speakers of different genders, ages, and ethnicities. It also covers formal (e.g., lectures, speeches, and debates) and casual (e.g., chats and Vlogs) scenarios. We adopt this dataset to train our audio-to-skeleton prediction model.

Our contributions can be summarized as follows. First, we propose a co-speech gesture video generation framework based on a lightweight audio-to-skeleton prediction model~\cite{super-resolution} and an off-the-shelf human video generation model. Second, to the best of our knowledge, we build the first publicly available large-scale dataset for co-speech gesture video generation. Third, extensive experiments demonstrate that our approach generates high-fidelity co-speech gesture videos across various speakers and scenarios, reliably promoting the performance of both general human video generation~\cite{stableanimator,mimicmotion} and co-speech gesture video generation models~\cite{echomimicv2}. We hope our work can expedite the research on high-fidelity co-speech gesture video generation.




\section{Related Works}
\label{sec:related}

\begin{figure*}[h]
    \centering
    \includegraphics[width=\linewidth]{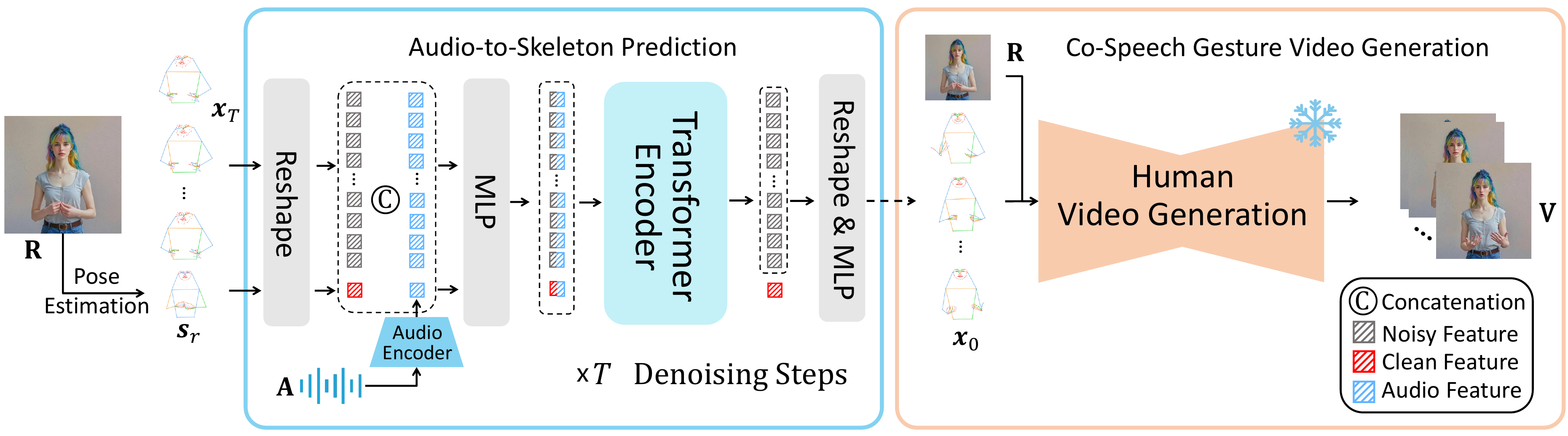}
    \vspace*{-5mm}
    \caption{Overview of our co-speech gesture video generation framework. We concatenate the 2D skeleton of the reference image $\textbf{R}$ with the noisy skeleton sequence $\textbf{\textit{x}}_T$ along the frame dimension, providing the body shape cue of the speaker. We then concatenate the embeddings of skeletons and those of audio segments along the feature dimension as the input of the diffusion model, enforcing strict temporal synchronization. Finally, we employ one off-the-shelf human video generation model to produce the co-speech gesture video $\textbf{V}$ with the synthesized skeleton sequence as an auxiliary condition.}
    \label{fig:model}
    \vspace*{-5mm}
\end{figure*}

\paragraph{Co-Speech Gesture Video Generation.} 
Existing works for co-speech gesture video generation can be categorized into the warping-based~\cite{angie,S2G-MDDiffusion,diffted} and warping-free~\cite{vlogger,cyberhost} methods. The warping-based methods first predict motion features according to the given audio and the speaker's reference image. These features are then utilized to warp the reference image into co-speech gesture video frames. According to the property of motion features, the warping-based methods can be further divided into implicit and explicit approaches. The implicit ones~\cite{angie} employ an unsupervised method~\cite{mraa} to model the latent motion features, while the explicit ones~\cite{S2G-MDDiffusion,diffted} usually adopt thin-plate spline (TPS) keypoints~\cite{tps,tps_transformation} to represent the motion features. The motion features are typically first transformed into optical flows~\cite{fomm}, by which the reference image can be warped into video frames. The downside of the warping operation is that it is prone to produce artifacts when encountering large pose variations. 

Several recent works~\cite{vlogger,cyberhost,echomimicv2} discard the warping operation. They are typically diffusion model-based, and their models are trained using private large-scale datasets. Besides the speaker’s reference image and the given audio, they generally adopt auxiliary conditions to achieve more fine-grained control. For example, Corona et al.~\cite{vlogger} predicted the sequence of 3D human motions~\cite{smpl} from audio as the auxiliary condition. Meng et al.~\cite{echomimicv2} adopted the sequence of hand skeletons extracted from another video. However, these auxiliary conditions may not be precise. Specifically, predicting the 3D motion itself is a challenging problem, while the sequence of hand skeletons extracted from another video can hardly match the target speaker’s audio and body shape.

Besides, existing public datasets for co-speech gesture video generation are usually small, with low image resolutions. For example, existing works~\cite{angie,S2G-MDDiffusion,diffted} only adopt the 1,200 clips of only four speakers and 1,322 cilps of TED talk shows selected from the PATS~\cite{pat2} and TED-talks~\cite{ted} datasets, respectively. Their image resolutions are 256$\times$256 and 384$\times$384 pixels, respectively. These datasets' limited scale and diversity have significantly affected the fidelity of co-speech gesture video generation. 

In this paper, we democratize high-fidelity co-speech gesture video generation by an expressive and reliable audio-to-skeleton prediction model and making use of off-the-shelf powerful human video generation models. We also contribute a large-scale dataset that surpasses existing public ones in scale, diversity, and quality.

\paragraph{Human Video Generation.} 
Human video generation~\cite{animateanyone,magicanimate,emo,echomimic,dreampose} aims to create realistic human videos from a given human image and other conditions such as audio and pose sequences. It includes many popular subtasks, e.g., full-body human video generation~\cite{animateanyone,magicanimate,dreampose} and talking face~\cite{emo,echomimic,sadtalker,aniportrait,follow-your-emoji} generation. The former~\cite{animateanyone,magicanimate,mimicmotion} usually adopts a sequence of skeleton~\cite{openpose,dwpose} as condition to produce full-body human videos. For example, Li et al. proposed the Animate Anyone method~\cite{animateanyone} that is based on the powerful Stable Diffusion model~\cite{stablediffusion}. It also introduces a reference net to Stable Diffusion, which further improves the appearance consistency between the obtained video frames and the reference human image. To promote model performance across different body shapes, recent works incorporate human body information, e.g., densepose~\cite{densepose,magicanimate}, 3D meshes~\cite{talk-act,make-your-anchor}, depth maps~\cite{champ}, and normal maps~\cite{smpl,smplx,champ}, as auxiliary conditions to the diffusion model. Besides, Zhang et al.~\cite{mimicmotion} proposed to incorporate keypoint detection confidence into the pose guidance, which enhances the model's robustness to keypoint detection errors. Finally, Tu et al.~\cite{stableanimator} proposed to utilize a face encoder to extract facial features from the reference image, which are then fed into the video generation model to enhance the fidelity of synthesized human videos.

Talking face generation~\cite{emo,echomimic,sadtalker,aniportrait,follow-your-emoji,letstalk} aims to create videos with natural facial expressions from a speech audio and a reference facial image. Existing methods can be categorized into one- and two-stage approaches. The one-stage methods~\cite{emo} adopt a diffusion model~\cite{stablediffusion} to directly synthesize talking face videos from the audio. In contrast, the two-stage methods~\cite{aniportrait,follow-your-emoji,echomimic} first generate intermediate face representations, e.g., 3D Morphable Model (3DMM) coefficients~\cite{3dmm}, and then feed them into a video generation model. Compared to this task, co-speech gesture video generation involves body parts below the shoulders.
\section{Methods}
\label{sec:method}

We first describe our co-speech gesture video generation framework in Section~\ref{sec:framework} and Figure~\ref{fig:model}. Then, we introduce the way to construct our large-scale CSG-405 database in Section~\ref{sec:dataset}.

\begin{figure*}[h]
    \centering
    \includegraphics[width=\linewidth]{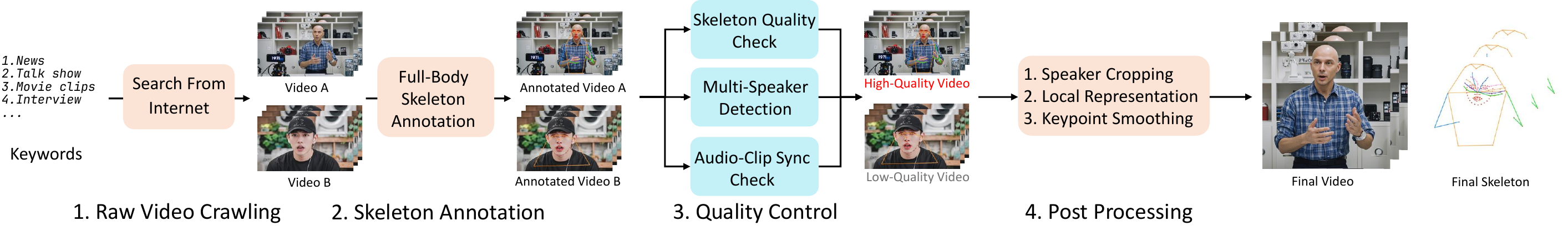}
    \vspace*{-5mm}
    
    \caption{An overview of our data collection pipeline.}
    \label{fig:data collection pipeline}
    \vspace*{-4mm}
    
\end{figure*}

\subsection{Co-Speech Gesture Video Generation}
\label{sec:framework}
\paragraph{Preliminaries.}We denote the audio as $\textbf{A}$, the reference image of the target speaker as $\textbf{R}$, and the skeleton extracted from $\textbf{R}$ as $\textbf{\textit{s}}_r \in \mathbb{R}^{1\times2K}$, where $K$ is the number of 2D body keypoints in the skeleton. Given a random pair of $\textbf{A}$ and $\textbf{\textit{s}}_r$, our audio-to-skeleton prediction model $G_s$ produces a sequence of high-fidelity and expressive skeletons $\textbf{S}_g \in \mathbb{R}^{F\times2K}$, where $F$ means the frame number. We feed $\textbf{S}_g$ and $\textbf{R}$ into an off-the-shelf human video generation model $G_v$, which produces the final co-speech gesture video $\textbf{V}$. The whole framework can be formulated as: $\textbf{S}_g=G_s(\textbf{\textit{s}}_r,\textbf{A})$ and $\textbf{V}=G_v(\textbf{S}_g,\textbf{R})$.

\paragraph{Audio-to-Skeleton Prediction.}We build $G_s$ as an audio-conditioned diffusion model, which synthesizes $\textbf{S}_g$ by progressive denoising from random Gaussian noise. During training, keypoint coordinates in each ground-truth skeleton sequence is represented as $\textbf{\textit{x}}_0\in \mathbb{R}^{F\times2K}$. After $T$ diffusion steps, $\textbf{\textit{x}}_0$ becomes an isotropic Gaussian noise $\textbf{\textit{x}}_T$. Then, we optimize parameters of $G_s$ by denoising $\textbf{\textit{x}}_T$ according to the time step $t$ and audio condition $\textbf{A}$. Following ~\cite{mdm}, we require $G_s$ to predict $\textbf{\textit{x}}_0$ instead of the added noise in each denoising step. The loss function can be formulated as: \[ L = \mathbb{E}_{\textbf{\textit{x}}_0, t} \left[ \| \textbf{\textit{x}}_0 - G_s(\textbf{\textit{x}}_t, t, \textbf{A}) \|_2^2 \right], \] where $\textit{\textbf{x}}_t$ denotes the noisy skeleton sequence in step $t$.  

As shown in Figure~\ref{fig:model}, we implement the above model according to the Diffusion Transformer (DiT) architecture~\cite{dit}. Moreover, as any artifact in the predicted skeleton harms the subsequent video generation process, we propose the following two strategies to achieve high-fidelity audio-to-skeleton prediction. 

First, we enable model prediction to accommodate the body shape of the target speaker. To achieve this goal, we concatenate $\textbf{\textit{x}}_t$ with $\textbf{\textit{s}}_r$ along the frame dimension as the input of the diffusion model for each denoising step $t$. By modeling their temporal dependencies with the self-attention layers, $G_s$ effectively leverages cues from $\textbf{\textit{s}}_r$ to generate $\textbf{S}_g$ that conforms to the speaker’s body shape. 

Second, we enforce strict coordination between $\textbf{S}_g$ and $\textbf{A}$. Existing approaches to co-speech gesture generation usually impose the audio condition via cross-attention~\cite{angie,S2G-MDDiffusion,probtalk}. In this scheme, the correspondence between each gesture and audio segment is obscure, harming the obtained gestures' fidelity. To handle this problem, we extract a sequence of more fine-grained segment-level audio features $\hat{\textbf{A}} \in \mathbb{R}^{F\times768}$ from $\textbf{A}$ using the wav2vec 2.0 model~\cite{wav2vec2}, where the segments number is the same as the video frame number. Then, we concatenate $\textbf{\textit{x}}_t$ with $\hat{\textbf{A}}$ along the feature dimension in each denoising step. To maintain dimension consistency between input embeddings, we concatenate a zero vector to $\textbf{s}_r$. Finally, we utilize a linear projection layer to project the concatenated features to a dimension suitable to the DiT layers. Compared to cross-attention~\cite{transformer}, our strategy ensures one-to-one correspondence between each gesture and audio segment, which promotes the fidelity and expressiveness of generated skeletons. 

Finally, we train $G_s$ with the classifier-free guidance strategy~\cite{cfg}. Specifically, we replace the audio condition $\textbf{A}$ with $\varnothing$ using a probability of 10\% during training, which results in an unconditional generation. When $\varnothing$ is adopted, we concatenate a set of learnable vectors that are randomly initialized to $\textbf{\textit{x}}_t$. During inference, we make a trade-off between generation diversity and fidelity as follows:\[
G_{s}'(\textbf{\textit{x}}_t, t, \textbf{A}) = G_s(\textbf{\textit{x}}_t, t, \varnothing) + \alpha \cdot (G_s(\textbf{\textit{x}}_t, t, \textbf{A}) - G_s(\textbf{\textit{x}}_t, t, \varnothing)),
\]
where $\alpha$ denotes the classifier-free guidance scale.

\paragraph{Co-Speech Gesture Video Generation.} We feed both the predicted skeleton sequence $\textbf{S}_g$ and the reference image $\textbf{R}$ into an off-the-shelf human video generation model $G_v$, which generates the final co-speech gesture video $\textbf{V}$. Since $\textbf{S}_g$ well aligns with the audio condition $\textbf{A}$ and the reference image $\textbf{R}$, $G_v$ can produce more realistic human videos compared to those using skeletons extracted from other reference videos~\cite{echomimicv2,stableanimator,mimicmotion}.


\subsection{Dataset Construction}
\label{sec:dataset}
\paragraph{Dataset Description} To further democratize the research on high-fidelity co-speech gesture video generation, we build a large-scale database named CSG-405 and make it publicly available. This dataset contains 147,550 clips that cover 71 common types of speeches, with a total duration of 405 hours. Each clip is of 5-15 seconds and adjusted to 25FPS. The mean duration of each clip is 9.88 seconds. We also provide high-quality skeleton annotation for each video frame. Each skeleton contains 133 keypoints on the face, hands, and other body parts. Besides, all audio signals are sampled at a rate of 16K. 

As shown in Table~\ref{tab:dataset comp} and Figure~\ref{fig:keypoints and dataset attribute}, our database features a large number of speakers. They are of different genders, ethnicities, ages, and emotions. Similar to~\cite{celeba_hq, celebv_hq}, the number of speaker identities and person attributes is estimated using a facial image understanding model~\cite{deepface}.

\paragraph{Data Collection Pipeline.} We illustrate our data collection pipeline in Figure~\ref{fig:data collection pipeline}. It incorporates four stages including raw video crawling, skeleton annotation, quality control, and post-processing.

First, we utilize GPT-4o~\cite{gpt-4o} to enumerate speech types. We search for videos of each speech type according to keyword from Youtube and download them with a high-resolution of 1280$\times$720 pixels. Figure~\ref{fig:keypoints and dataset attribute} (a) summarizes the proportion of obtained videos for each speech type, which indicates that our database covers diverse real-life speech types. 

Second, we annotate a full-body skeleton for each video frame. Specifically, we leverage the powerful DWPose model~\cite{dwpose} to annotate 133 body keypoints and also obtain a confidence score for each keypoint. The obtained keypoints follow the COCO-Whole-Body format~\cite{coco-wholebody}. As shot transitions may occur within a video, we detect sudden changes in camera motion or image color. According to the detected changes, we segment each video into clips with a length of 5-15 seconds. 

Third, we carefully check the quality of skeletons, video, and audio. For the skeleton, we remove clips with missing upper body keypoints, those captured from the side or back viewpoints, those containing very small human figures, and those with basically static poses. If multiple persons appear in a video, we process each person’s skeleton independently. For the video and audio, we utilize PyAnnote~\cite{pyannote} to detect and remove clips where multiple persons speak simultaneously. We also adopt SyncNet~\cite{syncnet} to remove clips where the lip movements misalign with the audio.

Finally, we crop each speaker with margins from video frames and resize all cropped regions to 512×512 pixels. We also convert keypoint coordinates into local motion representations~\cite{sdt}. Specifically, we adopt the nose tip and wrists as the root nodes for keypoints on the face and hands, respectively. For keypoints on the other body parts, we utilize the neck keypoint as the root node. Then, we compute the coordinate of each keypoint relative to its corresponding root node. This strategy disentangles holistic motion with subtle facial and hand movements, facilitating our model to synthesize high-fidelity and expressive skeletons. Besides, we apply temporal smoothing on all keypoints except for those on the mouth to reduce irregular jittering. As the mouth usually moves drastically during speech, we keep keypoint coordinates on the mouth unchanged.

\section{Experiments}
\label{sec:exp}

\begin{table*}
\centering
    \caption{Quantitative comparisons between our method and state-of-the-art methods on PATS~\cite{pat2}, TED-talks~\cite{ted}, and our CSG-405 datasets. Ours$^\dagger$ (StableAnimator) means that we fine-tune StableAnimator on our CSG-405 database.}
    \vspace*{-4mm}
    \scriptsize  
\label{tab_quantative}
\resizebox{\textwidth}{!}{
    \begin{tabular}{llccccccc}
    \toprule
    \multicolumn{1}{c}{\multirow{2}*{Dataset}} & 
    \multicolumn{1}{c}{\multirow{2}*{Method}} & \multicolumn{2}{c}{Paired} & \multicolumn{5}{c}{Unpaired} \\
    \cmidrule(lr){3-4} \cmidrule(lr){5-9}
    \multicolumn{2}{c}{} & {SSIM$\uparrow$} & {PSNR$\uparrow$} & {CSIM$\uparrow$} & {FID$\downarrow$} & {FVD$\downarrow$} & {Sync-C$\uparrow$} & {Sync-D$\downarrow$} \\

    \midrule
    {} & S2G-MDDiffusion~\cite{S2G-MDDiffusion} & 0.59 & 16.84 & \textbf{0.77} & 85.23 & 989.98 & 3.94 & 10.47 \\
    {} & EchoMimicV2~\cite{echomimicv2} &  0.57 & 16.35 & 0.73 & 97.86 &  1243.48 & 5.44 & 9.01 \\
    {PATS} & StableAnimator~\cite{stableanimator} &  \textbf{0.75} & \textbf{22.93} & 0.63 & 91.83 & 1131.73  & 4.05 & 9.88 \\
    {} & Ours (EchoMimicV2) & 0.56 & 15.70 & 0.73 & 89.36 & 1113.50 & 5.41 & \textbf{9.00} \\
    {} & Ours (StableAnimator) & 0.56 & 16.35 & 0.76 & 73.56 & 928.26 & 5.22 & 9.72 \\
    {} & Ours$^\dagger$ (StableAnimator) & 0.60 & 16.68 & \textbf{0.77} & \textbf{70.38} & \textbf{877.07} & \textbf{5.49} & 9.36 \\
    \midrule  
    
    {} & DiffTED~\cite{diffted}  &  0.71 & 18.53 & 0.75 & 83.80 & 1138.29 & 0.74 & 13.05 \\
    {} & EchoMimicV2~\cite{echomimicv2} &  0.66 & 18.09 & 0.71 & 86.88 & 943.08 & 4.58 & 9.65 \\
    {TED-talks} & StableAnimator~\cite{stableanimator} &  \textbf{0.81} & \textbf{25.50} & 0.55 & 88.29 &  890.48  & 4.10 & 10.20 \\
    {} & Ours (EchoMimicV2) & 0.65 & 17.67 & 0.74 & 77.46 & 785.47 & 4.88 & 9.41 \\
    {} & Ours (StableAnimator) & 0.61 & 17.18 & 0.75 & \textbf{68.84} & 670.21 & 5.44 & 9.21 \\
    {} & Ours$^\dagger$ (StableAnimator) & 0.63 & 17.30 & \textbf{0.76} & 70.05 & \textbf{650.20} & \textbf{5.59} & \textbf{9.15} \\
    \midrule  

    {} & S2G-MDDiffusion~\cite{S2G-MDDiffusion}  &  0.55 & 14.02 & 0.39 & 216.80 & 2538.70 & 2.48 & 11.54 \\
    {} & DiffTED~\cite{diffted}  &  0.67 & 15.98 & 0.61 & 138.82 &  1852.10 & 0.66 & 13.06 \\
    {} & MimicMotion~\cite{mimicmotion} &  0.75 & 20.47 & 0.60 & 115.65 & 1751.33 & 3.26 & 11.12 \\
    {} & EchoMimicV2~\cite{echomimicv2} &  0.71 & 17.88 & \textbf{0.82} & 90.01 & 1444.00 & 6.53 & 8.09 \\
    {CSG-405} & StableAnimator~\cite{stableanimator} &  \textbf{0.83} & \textbf{24.27} & 0.67 & 94.94 & 1532.74  & 5.13 & 9.50 \\
    {} & Ours (MimicMotion) & 0.59 & 14.98 & 0.76 & 73.77 & 1264.22 & 4.43 & 10.36 \\
    {} & Ours (EchoMimicV2) & 0.69 & 17.02 & \textbf{0.82} & 70.03 & 1111.61 & \textbf{6.59} & \textbf{8.04} \\
    {} & Ours (StableAnimator) & 0.65 & 16.26 & 0.81 & 66.47 & 988.43 & 5.82 & 9.11 \\
    {} & Ours$^\dagger$ (StableAnimator) & 0.67 & 16.62 & \textbf{0.82} & \textbf{62.93} & \textbf{984.13} & 6.28 & 8.68 \\
    
    \bottomrule
    \end{tabular}
    }
\vspace*{-4mm}
\end{table*}

\paragraph{Datasets and Metrics.}We conduct experiments on PATS~\cite{pat2}, TED-talks~\cite{ted}, and our collected CSG-405 datasets. We randomly select 133 clips from our CSG-405 database for testing and divide the remaining data into 90\% for training and 10\% for validation. The training, validation and testing data are about 364, 41, and 0.5 hours, respectively. After training on CSG-405, the obtained models are directly evaluated on PATS and TED-talks without further fine-tuning. Following~\cite{angie,S2G-MDDiffusion}, we employ the 474 clips of four speakers (Jon, Kubinec, Oliver, and Seth) from the PATS~\cite{pat2} database for testing. The mean length of these clips is 9.8 seconds and the resolution of video frames is unified to 256$\times$256 pixels. The testing set of the TED-talks~\cite{ted} database include 122 clips. The resolution of video frames is unified to 384$\times$384 pixels and the frame number ranges from 64 to 1,024 in each clip.

We perform evaluations using both the paired and unpaired configurations. In the former setting, the audio and reference image are from the same video, while the latter composes audio-image pairs from different videos. Existing methods~\cite{cyberhost,echomimicv2,vlogger} typically overlook the latter setting, which is evidently more practical in real-world applications. For the paired setting, we adopt Structural Similarity (SSIM)~\cite{ssim} and Peak Signal to Noise Ratio (PSNR)~\cite{psnr} to evaluate the visual quality of the generated videos. For the unpaired setting, we employ Fréchet Inception Distance (FID)~\cite{fid} and Fréchet Video Distance (FVD)~\cite{fvd} to evaluate the general video quality. Additionally, we employ SyncNet~\cite{syncnet} to calculate Sync-C and Sync-D, which assess audio-lip synchronization accuracy. We adopt the cosine similarity (CSIM) to evaluate the identity consistency.

\begin{figure*}[h]
    \centering
    \includegraphics[width=\linewidth]{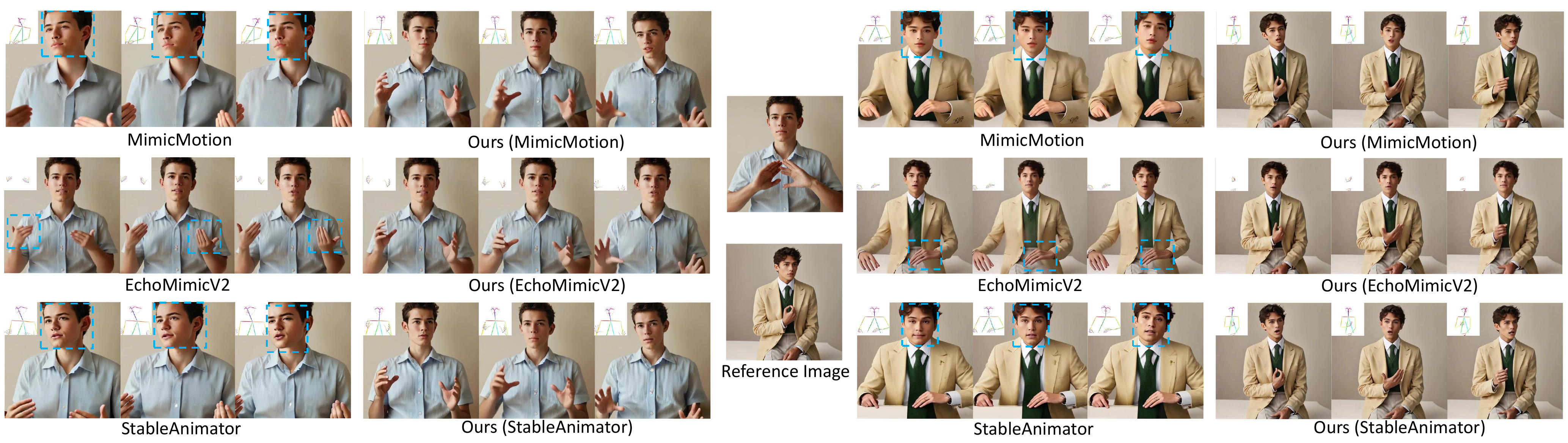}
    \vspace{-6mm}
    \caption{Qualitative comparisons between our methods and state-of-the-art methods on our CSG-405 database.}
    \label{fig:compare_on_ours}
    \vspace{-6mm}
\end{figure*}

\begin{figure*}[h]
    \centering
    \includegraphics[width=\linewidth]{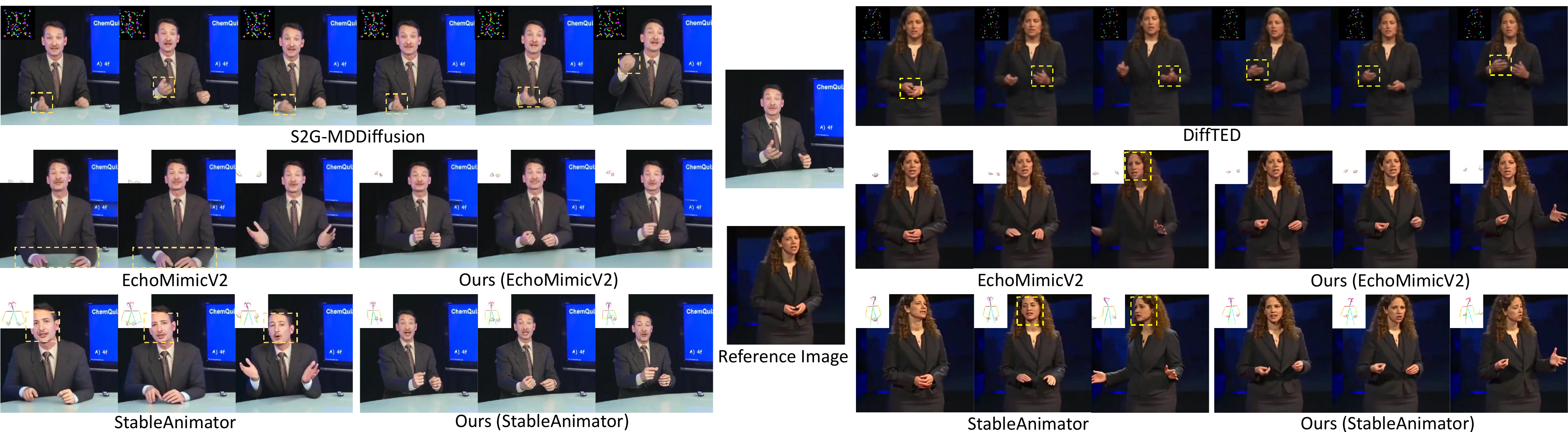}
    \vspace{-6mm}
    \caption{Qualitative comparisons between our methods and state-of-the-art methods on PATS~\cite{pat2} and TED-talks~\cite{ted} databases.}
    \label{fig:compare_on_ted_and_pats}
    \vspace{-6mm}
\end{figure*}

\paragraph{Implementation Details.}
We employ four V100 GPUs with a memory size of 32GB to train our audio-to-skeleton prediction model. We set the batch size to 128 and adopt the Adam optimizer~\cite{kingma2014adam} with a learning rate of 5e-5 to train this model for 2,000,000 steps. With the obtained skeleton sequences, we employ one general human video generation model named StableAnimator~\cite{stableanimator} and one co-speech gesture video generation model named EchoMimicV2~\cite{echomimicv2} for final video generation, respectively. They are denoted as ``Ours (StableAnimator)'' and ``Ours (EchoMimicV2)'' in Figure~\ref{fig:compare_on_ours}, Figure~\ref{fig:compare_on_ted_and_pats}, and Table~\ref{tab_quantative}. It is worth noting that EchoMimicV2 adopts the hand skeletons rather than full-body skeletons as auxiliary control condition, so we extract the hand skeletons from our produced full-body skeletons for ``Ours (EchoMimicV2)''. For evaluations on our CSG-405 database, we further employ one general human video generation method named MimicMotion~\cite{mimicmotion} to evaluate the generalizability of our audio-to-skeleton prediction model, which is denoted as ``Ours (MimicMotion)''.

\begin{figure*}[h]
    \centering
    \includegraphics[width=\linewidth]{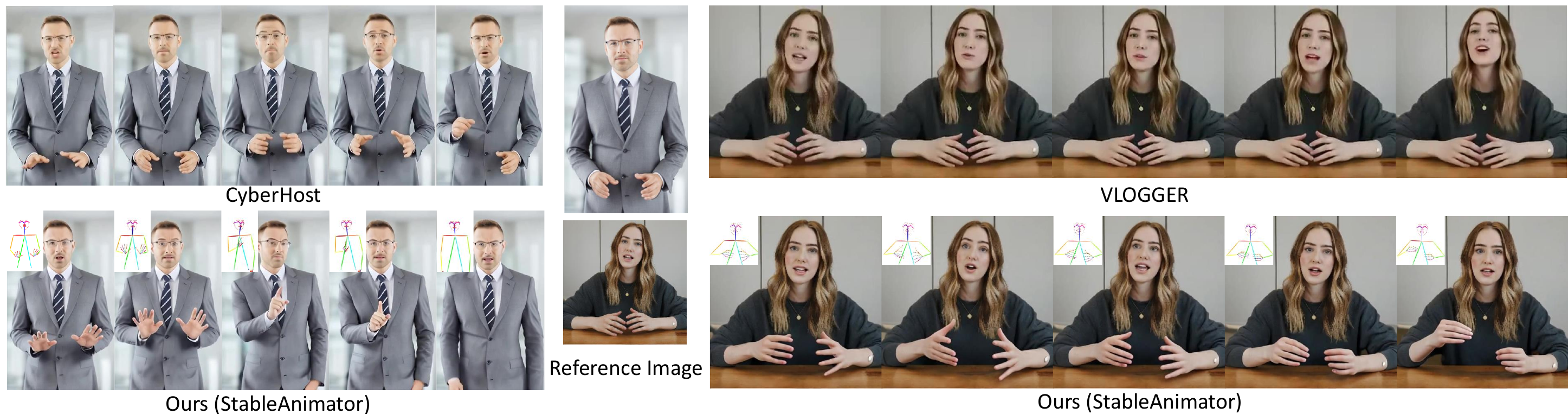}
    \vspace{-7mm}
    \caption{Qualitative comparisons between CyberHost~\cite{cyberhost}, VLOGGER~\cite{vlogger}, and our approach.}

    \label{fig:compare_on_cyberhost_vlogger}
    \vspace{-4mm}
\end{figure*}

\subsection{Qualitative Comparisons}
We perform qualitative comparisons between our approaches and state-of-the-art methods in Figure~\ref{fig:compare_on_ours}, Figure~\ref{fig:compare_on_ted_and_pats} and Figure~\ref{fig:compare_on_cyberhost_vlogger}. The methods in comparison include co-speech gesture generation methods such as S2G-MDDiffusion~\cite{S2G-MDDiffusion}, DiffTED~\cite{diffted}, EchoMimicV2~\cite{echomimicv2}, VLOGGER~\cite{vlogger} and CyberHost~\cite{cyberhost}, as well as general full-body human video generation methods like MimicMotion~\cite{mimicmotion} and StableAnimator~\cite{stableanimator}. 

First, it is shown that the warping-based methods~\cite{S2G-MDDiffusion,diffted} tend to produce blurry body parts, e.g., hands, when encountering large pose changes, as highlighted in Figure~\ref{fig:compare_on_ted_and_pats}. In contrast, our method is warping-free and able to produce clearer body parts. Second, Figure~\ref{fig:compare_on_ours} and Figure~\ref{fig:compare_on_ted_and_pats} show that methods that require driven skeleton condition, e.g., MimicMotion~\cite{mimicmotion}, StableAnimator~\cite{stableanimator}, and EchoMimicV2~\cite{echomimicv2}, may produce body shapes that contradict with those of the reference images in the unpaired setting. This is because the driven skeleton condition they adopt is extracted from another video. In comparison, our method significantly reduces this error by generating skeletons according to the body shape revealed in the reference image. 

Since the VLOGGER~\cite{vlogger} and CyberHost~\cite{cyberhost} models are not publicly available, we perform qualitative comparisons with them using demo videos from their respective project homepages. As shown in Figure~\ref{fig:compare_on_cyberhost_vlogger}, our approach surpasses both VLOGGER and CyberHost in the naturalness and diversity of synthesized gestures.




\subsection{Quantitative Comparisons}
We further present quantitative comparisons in Table~\ref{tab_quantative}. Our methods achieve the best overall performance across the PATS, TED-talks, and CSG-405 datasets in the unpaired setting. 

As presented in the table, existing human video generation methods, e.g., StableAnimator~\cite{stableanimator}, demonstrate strong performance in the paired setting, suggesting their strong potential for co-speech gesture video generation. However, these approaches exhibit significant degradation under unpaired conditions, which is primarily due to the misalignment between the driven skeleton condition and the reference image. Our methods address this limitation by adaptive skeleton generation according to the body shape of the reference image, achieving state-of-the-art results in unpaired scenarios. In the paired setting, StableAnimator achieves the best SSIM and PSNR metrics. This is because they directly utilize ground-truth skeletons as a condition, while we generate skeletons with randomness from audio. It is also worth noting that EchoMimicV2 achieves suboptimal performance in paired settings. This is because it tends to generate videos with the speaker facing forward and positioned strictly centrally, which can be misaligned with the original co-speech gesture video. We also conduct experiments using the StableAnimator finetuned on our CSG-405 database and observed improved performance. This indicates that our database can further enhance the performance of existing human video generation models for co-speech gesture video generation.


    



\begin{table}
\centering
    \caption{Ablation study on each key component of our model on CSG-405 dataset.}
    \vspace{-3mm}
\label{tab_ablation}
\resizebox{0.5\textwidth}{!}{
    \begin{tabular}{lcccccccc}
    \toprule

    {Method} & {SSIM$\uparrow$} & {PSNR$\uparrow$}& {CSIM$\uparrow$} & {FID$\downarrow$} & {FVD$\downarrow$} &{Sync-C$\uparrow$} & {Sync-D$\downarrow$}  \\
    
    \midrule

    w/o Reference Skeleton  & 0.57 & 13.75 & 0.57 &  127.40 & 1704.49 & 4.88 & 9.94  \\
    Cross-Attention Conditioning & 0.63 & 15.71 & 0.76 & 78.10 & 1023.52  & 0.77 & 13.64  \\
    w/o Local Representation & 0.62 & 15.02 & 0.71 &  86.69 & 1073.47 & 0.49 & 13.59 \\
    \textbf{Ours} & \textbf{0.67} & \textbf{16.62} &\textbf{ 0.82} & \textbf{62.93} & \textbf{984.13} & \textbf{6.28}  & \textbf{8.68}  \\

    \bottomrule
    \end{tabular}
}
\vspace*{-7mm}
\end{table}

\subsection{Ablation Study}
Finally, we conduct ablation study on our CSG-405 database to justify each key design in our audio-to-skeleton prediction model. We adopt the finetuned StableAnimator as the video generator and summarize experimental results in Table~\ref{tab_ablation}.

\paragraph{Effectiveness of the Reference Skeleton.}In this experiment, we evaluate whether it is beneficial to adopt the reference skeleton $\textbf{\textit{s}}_r$ as the input of our audio-to-skeleton prediction model. This experiment is denoted as ``w/o Reference Skeleton'' in Table~\ref{tab_ablation}. It is shown that its performance is significantly lower than our audio-to-skeleton prediction model. This is reasonable as it loses the prior about the target speaker’s body shape reflected in the reference image. Therefore, it tends to produce skeletons with incorrect body shapes, resulting in unrealistic co-speech gesture videos.

\paragraph{Effectiveness of Our Audio Conditioning.}As described in Section~\ref{sec:framework}, we concatenate the embeddings of the skeleton and audio segments along the feature dimension as the input of the diffusion model. In this experiment, we justify that this audio conditioning strategy outperforms the popular cross-attention method~\cite{transformer}. As shown in Table~\ref{tab_ablation}, our audio conditioning mechanism achieves better Sync-C and Sync-D metrics, indicating better performance in synchronization between the audio condition and synthesized gestures. This is because compared with the common cross-attention strategy, our approach enforces fine-grained correspondence between the audio and skeleton segments.

\paragraph{Effectiveness of Local Motion Representation.}As introduced in Section~\ref{sec:dataset}, we adopt the local motion representation~\cite{sdt} for keypoints on the face and hands in our model. As revealed in Table~\ref{tab_ablation}, the local motion representation leads to more realistic videos with better audio-gesture synchronization. The reason is that it disentangles the global skeleton motion with the subtle facial and hands movements, enabling our model focus more on generating expressive facial expressions and hand movements.

\vspace{-3mm}
\section{Conclusion}
\label{sec:conclusion}
In this paper, we present a novel co-speech gesture video generation framework based on audio-conditioned skeleton prediction. Our method bridges the audio-visual alignment challenge by integrating reference-aware skeleton generation with feature-level audio-skeleton fusion. Furthermore, we introduce CSG-405, the first large-scale public dataset with diverse speech scenarios and high-quality annotations, which addresses the data scarcity problem in existing research. Experiments demonstrate that our approach significantly outperforms current state-of-the-art methods in both video quality and audio-visual synchronization. By releasing both the technical framework and dataset, we aim to democratize high-fidelity co-speech gesture synthesis.
~\paragraph{Broader Impacts.}This work democratizes high-fidelity co-speech gesture video generation by proposing a audio-to-skeleton prediction model and a large-scale dataset. While our work advances co-speech gesture video synthesis for human-centered applications, we advocate for the development of ethical frameworks (e.g., synthetic content detection tools) alongside generative technologies to ensure their responsible deployment and societal well-being.
~\paragraph{Acknowledgement.}This work was supported by the 2024 Tencent AI Lab Rhino-Bird Focused Research Program, the National Natural Science Foundation of China under Grant 62476099 and 62076101, Guangdong Basic and Applied Basic Research Foundation under Grant 2024B1515020082 and 2023A1515010007, the Guangdong Provincial Key Laboratory of Human Digital Twin under Grant 2022B1212010004, and the TCL Young Scholars Program.




\end{document}